\documentclass[10pt,twocolumn,letterpaper]{article}

\usepackage[pagenumbers]{cvpr} 
\usepackage{float}
\usepackage{booktabs}
\usepackage{multirow}
\usepackage[accsupp]{axessibility}

\definecolor{cvprblue}{rgb}{0.21,0.49,0.74}
\usepackage[pagebackref,breaklinks,colorlinks,allcolors=cvprblue]{hyperref}

\def\paperID{19}
\def\confName{CVPR}
\def\confYear{2026}

\title{IndoorCrowd: A Multi-Scene Dataset for Human Detection, Segmentation, and Tracking with an Automated Annotation Pipeline}

\author{Sebastian-Ion Nae$^1$, \ Radu Moldoveanu$^1,^2$, \ Alexandra Stefania Ghita$^1$, \ Adina Magda Florea$^1$ \\
$^1$National University of Science and Technology Politehnica Bucharest, Romania \\
$^2$Expleo, Romania \\
{\tt\small sebastian\_ion.nae@stud.fils.upb.ro \ radu.moldoveanu2112@stud.electro.upb.ro} \\
{\tt\small stefania.a.ghita@upb.ro \ adina.florea@upb.ro}
}

\begin{document}

\maketitle

\begin{abstract}
Understanding human behaviour in crowded indoor environments is central to surveillance, smart buildings, and human-robot interaction, yet existing datasets rarely capture real-world indoor complexity at scale. We introduce IndoorCrowd, a multi-scene dataset for indoor human detection, instance segmentation, and multi-object tracking, collected across four campus locations (ACS-EC, ACS-EG, IE-Central, R-Central). It comprises $31$ videos ($9{,}913$ frames at $5$\,fps) with human-verified, per-instance segmentation masks. A $620$-frame control subset benchmarks three foundation-model auto-annotators: SAM3~\cite{carion2026sam3}, GroundingSAM~\cite{ren2024grounded}, and EfficientGroundingSAM~\cite{xiong2023efficientsam}, against human labels using Cohen's $\kappa$, AP, precision, recall, and mask IoU. A further $2{,}552$-frame subset supports multi-object tracking with continuous identity tracks in MOTChallenge format. We establish detection, segmentation, and tracking baselines using YOLOv8n~\cite{sohan2024review}, YOLOv26n~\cite{sapkota2025yolo26}, and RT-DETR-L~\cite{zhao2024detrs} paired with ByteTrack~\cite{zhang2022bytetrack}, BoT-SORT~\cite{aharon2022bot}, and OC-SORT~\cite{cao2023observation}. Per-scene analysis reveals substantial difficulty variation driven by crowd density, scale, and occlusion: ACS-EC, with $79.3\%$ dense frames and a mean instance scale of $60.8$\,px, is the most challenging scene. The project page is available at \url{https://sheepseb.github.io/IndoorCrowd/}.
\end{abstract}
\section{Introduction}
\label{sec:intro}
Detecting and tracking people in indoor spaces is a foundational task~\cite{awais2025foundation} for crowd management~\cite{Ong_2023_ICCV, he2025learning}, response planning~\cite{HUANG2023106285} and human-robot~\cite{chen2019crowd}. Despite substantial progress driven by large-scale outdoors benchmarks such as CrowdHuman~\cite{shao2018crowdhuman}, WiderPerson~\cite{zhang2019widerperson} and the MOTChallenge series~\cite{dendorfer2021motchallenge}, indoor environments remain severely underrepresented. Outdoor datasets are dominated by street-level or vehicle-centric viewpoints, which exhibit different density distributions, illumination profiles and occlusion patterns than indoors corridors, atria or entrance halls. Those environments introduce a new set of challenges, camera fields of view are obstructed by pillars, furniture and architectural features, producing frequent inter-person occlusions at IoU levels that can no longer use standard NMS. Crowd density fluctuates sharply within short temporal windows~\cite{he2025learning}, between nearly empty corridors and congested atria, creating wide intra-scene variance that stresses both sparse and dense detection regimes. A further barrier is the annotation cost. Instance-level mask labels are substantially more expensive than bounding boxes, while tracking is even more so due to the required temporally consistent identity assignment~\cite{zhang2025efficiently}. Recent foundation models such as SAM~\cite{kirillov2023segment}, GroundingDINO~\cite{liu2024grounding} and their derivatives, offer the possibility of automated pre-annotation that humans can verify rather than produce from scratch~\cite{barsellotti2025talking, nae2026learningflyreplaybasedcontinual}. However, the quality of these auto-labels in crowded indoor scenes and their suitability as a basis for human correction have not been characterised for new environments.

We present a new multi-scene indoor dataset for human detection, instance segmentation and multi-object tracking, collected across four distinct locations within a campus: ACS-EC, ACS-EG, IE-Central and R-Central. Each scene represents a different architectural layout, camera angle and crowd density regime. We sampled $31$ videos at $5$ fps, yielding $9,913$ frames and produced two annotation subsets: (i) instance segmentation masks and bounding boxes, and (ii) a tracking set of $2,552$ frames with human-verified continuous identity tracks in MOTChallenge format.

We evaluate $3$ foundation-model annotators: SAM3~\cite{carion2026sam3}, GroundingSAM~\cite{ren2024grounded} and EfficientGroundingSAM~\cite{xiong2023efficientsam} on all $620$ manually labelled frames using Cohen's $\kappa$~\cite{cohen1960coefficient, mchugh2012interrater}, AP@0.5, AP@0.75, precision, recall and mask IoU per scene. SAM3 achieves the highest recall ($0.88-0.98$) but low precision on dense scenes ($0.52$ in ACS-EC), making it a useful starting point for human correction. GroundingSAM variants offer better precision at the cost of moderate recall.

Our main contributions are: (i) a new indoor dataset with bounding box, instance segmentation and MOT tracking annotations across 4 diverse scenes; (ii) a study of auto-labelling quality across three foundation-models; (iii) a semi-automatic annotation pipeline combining high-recall auto-labelling with human correction and track curation; and (iv) detection, segmentation, and tracking baselines for future indoor human perception research. The dataset and annotations will be publicly released after acceptance.

\section{Related Work}
\label{sec:related_work}
\textbf{Pedestrian Detection Datasets} The Caltech Pedestrian Dataset~\cite{dollar_wojek_schiele_perona_2009} established early benchmarks for urban driving footage. CityPersons~\cite{zhang2017citypersons} broadened scene diversity, while CrowdHuman~\cite{shao2018crowdhuman} specifically targeted dense crowd scenarios with $15,000$ training images annotated with full-body, visible-body, and head boxes. WiderPerson~\cite{zhang2019widerperson} introduced further diversity across five categories. More recently, MMPD~\cite{zhang2024when} aggregated RGB, infrared, event-camera, and LiDAR datasets to benchmark multimodal pedestrian detection. However, these benchmarks are predominantly outdoor: street-level surveillance, intersections, or vehicle-centric views exhibiting lighting and density characteristics that differ fundamentally from indoor public spaces. Within the indoor domain, a crowd detection framework was proposed for surveillance, but it did not release a benchmark dataset. HRBUST-LLPED~\cite{li2023hrbust} addressed indoor low-light detection on wearable cameras. The JTA dataset~\cite{fabbri2018learning} provides large-scale indoor/outdoor tracking data synthetically. As shown in Table~\ref{tab:dataset_comparison}, our dataset combines real-world indoor capture, instance-level masks, and multi-object tracking annotations across diverse scenes, filling this niche.

\textbf{Multi-Object Tracking} The MOTChallenge series: MOT15~\cite{leal2015motchallenge}, MOT16, MOT17~\cite{milan2016mot16benchmarkmultiobjecttracking}, and MOT20~\cite{dendorfer2020mot20benchmarkmultiobject} has been the primary driver of progress in pedestrian tracking, providing standardised evaluation with MOTA, IDF1, ID-switch, MT, and ML metrics. MOT20 specifically targeted very dense crowd scenarios (up to $246$ persons/frame) in indoor and outdoor unconstrained environments~\cite{dendorfer2020mot20benchmarkmultiobject}. While these benchmarks focus on extreme crowding, they often lack instance masks; extensions like MOT20 provide segmentation-aware tracking, yet remain dominated by outdoor or controlled event footage. In contrast, JRDB-PanoTrack~\cite{le2024jrdb} provides large-scale indoor/outdoor panoptic tracking for robotics, but its robot-centric panoramic views differ from the fixed surveillance framing required for campus crowd management. Furthermore, while synthetic resources like MOTSynth~\cite{fabbri2021motsynth} enable segmentation and tracking at scale, they lack the nuanced real-world indoor dynamics, such as architectural occlusions and sharp density fluctuations, found in our dataset.

DanceTrack~\cite{sun2022dance} challenged trackers with non-linear motion and a similar appearance across identities. Despite these advances, MOTChallenge sequences do not fully reflect the variable geometry, density fluctuations, and field-of-view constraints of other interiors~\cite{du2024exploring}. The Cchead dataset~\cite{sun2025towards} offered crowd head tracking across classroom and outdoor scenes but focused exclusively on head-level bounding boxes.

\textbf{Foundation Models for Dataset Annotation} The Segment Anything Model (SAM)~\cite{kirillov2023sam} introduced a promptable with point segmentation architecture trained on $11$ billion masks, demonstrating strong zero-shot generalisation. SAM2~\cite{ren2024grounded} extended this to video through a streaming memory architecture, enabling temporally consistent mask propagation: the capability we exploit for human correction via SAM2.1. GroundingDINO~\cite{zhang2022dino} coupled a DINO-based~\cite{zhang2022dino} detector with language-grounded prompting for open-vocabulary detection. GroundingSAM~\cite{ren2024grounded} combines GroundingDINO with SAM to produce prompted instance masks without category-specific training. SAM3~\cite{carion2026sam3} further advances concept-driven via segmentation by using text prompts and achieves high recall. EfficientGroundingSAM reduces inference cost while preserving accuracy, as confirmed by our per-scene evaluation. We benchmark these auto-labellers in crowded indoor conditions using Cohen's $\kappa$ as a metric against human ground truth.

\begin{table}[t]
\centering
\caption{Comparison of pedestrian detection and tracking datasets. $\checkmark$: available; $\circ$: partial or indirect; \textemdash: not available. IndoorCrowd is one of the real-world indoor datasets providing all three annotation types across multiple scenes.}
\label{tab:dataset_comparison}

\resizebox{\columnwidth}{!}{%
\setlength{\tabcolsep}{3pt}
\begin{tabular}{l c c r r c c c c}

\toprule
\textbf{Dataset} & \textbf{Year} & \textbf{Indoor} 
  & \textbf{Frames} & \textbf{Scenes} 
  & \textbf{BBox} & \textbf{Mask} & \textbf{MOT} 
  & \textbf{Real} \\
\midrule

Caltech~\cite{dollar_wojek_schiele_perona_2009}   
  & 2009 & \textemdash & 250K  & 1  
  & $\checkmark$ & \textemdash & \textemdash & $\checkmark$ \\

MOT17~\cite{milan2016mot16benchmarkmultiobjecttracking} 
  & 2016 & \textemdash & 11K   & 7  
  & $\checkmark$ & \textemdash & $\checkmark$ & $\checkmark$ \\

CityPersons~\cite{zhang2017citypersons}           
  & 2017 & \textemdash & 35K   & 1  
  & $\checkmark$ & \textemdash & \textemdash & $\checkmark$ \\

CrowdHuman~\cite{shao2018crowdhuman}              
  & 2018 & \textemdash & 15K   & 1  
  & $\checkmark$ & \textemdash & \textemdash & $\checkmark$ \\

JTA~\cite{fabbri2018learning}                     
  & 2018 & $\circ$     & 460K  & 1  
  & $\checkmark$ & \textemdash & $\checkmark$ & \textemdash \\

WiderPerson~\cite{zhang2019widerperson}           
  & 2019 & \textemdash & 13K   & 1  
  & $\checkmark$ & \textemdash & \textemdash & $\checkmark$ \\

MOT20~\cite{dendorfer2020mot20benchmarkmultiobject}    
  & 2020 & $\circ$     & 13K   & 4  
  & $\checkmark$ & \textemdash & $\checkmark$ & $\checkmark$ \\

MOTSynth~\cite{fabbri2021motsynth}                
  & 2021 & \textemdash & 1.3M & 768  
  & $\checkmark$ & $\checkmark$ & $\checkmark$ & \textemdash \\

DanceTrack~\cite{sun2022dance}                    
  & 2022 & \textemdash & 105K  & 1  
  & $\checkmark$ & \textemdash & $\checkmark$ & $\checkmark$ \\

JRDB-PanoTrack~\cite{le2024jrdb}     
  & 2023 & $\checkmark$ & 1.9M & 27  
  & $\checkmark$ & \textemdash & $\checkmark$ & $\checkmark$ \\

HRBUST-LLPED~\cite{li2023hrbust}                  
  & 2023 & $\checkmark$ & 10K  & 1  
  & $\checkmark$ & \textemdash & \textemdash & $\checkmark$ \\

CcHead~\cite{sun2025towards}                      
  & 2025 & $\circ$     & 33K   & 2  
  & $\checkmark$ & \textemdash & $\checkmark$ & $\checkmark$ \\

\midrule
\textbf{IndoorCrowd (Ours)}                       
  & 2025 & $\checkmark$ & 9,913 & 4  
  & $\checkmark$ & $\checkmark$ & $\checkmark$ & $\checkmark$ \\

\bottomrule
\end{tabular}%
}
\end{table}

\textbf{Annotation Quality and Inter-Annotator Agreement.} Cohen's $\kappa$ is a standard measure of agreement between annotators, correcting for chance agreement~\cite{landis1977application}. Its application to object detection, computed over matched instance pairs at a fixed IoU threshold, provides a principled way to compare automated and human labels~\cite{tschirschwitz2025label}. Prior work on annotation quality in detection focused primarily on bounding-box ambiguity or label noise in large-scale datasets, with limited treatment of mask-level agreement in crowded scenes~\cite{ma2022effect, alhazmi2021effects}. We build an evaluation protocol for comparing multiple models against human ground truth across 4 indoor scenes, providing a reusable methodology for future dataset collection.

\begin{figure*}[t]
    \centering
    \includegraphics[width=\linewidth]{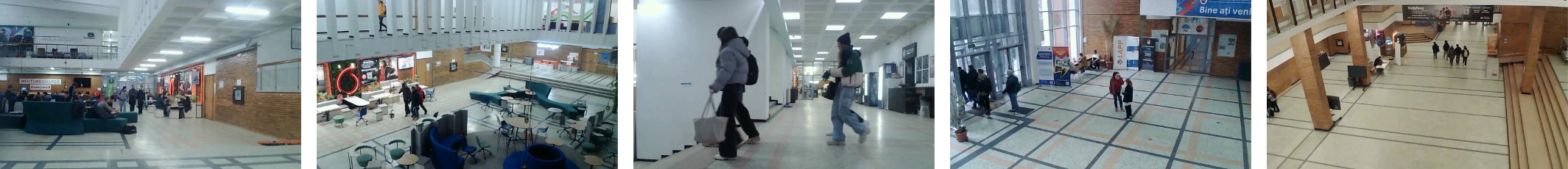}
    \caption{Representative frames from the four dataset scenes (left to right): ACS-EC ground-level view showing dense seating and circulation areas; ACS-EC elevated view providing a top-down perspective of the same atrium; ACS-EG narrow corridor with strong near-to-distal scale variance; IE-Central entrance hall captured from an elevated angle; and R-Central central atrium with prominent structural columns and overhead viewpoint.}
    \label{fig:samples}
\end{figure*}

\section{Data Collection}
\label{sec:data_collection}
We collected a dataset of surveillance-style videos across four distinct locations within a university campus. The dataset captures diverse challenges, including viewpoint variation, partial occlusion, and varying crowd density under naturalistic lighting. Data collection was approved by the university and conducted entirely in publicly accessible areas.

\subsection{Recording Setup}

Videos were captured using a fixed, mounted webcam at a resolution of $1280 \times 720$ pixels and a frame rate of $25$\,fps. All recordings were conducted during afternoon and evening hours on regular working days, capturing natural variation in crowd density and ambient lighting conditions ranging from bright artificial illumination to dimmer, mixed-light evening settings.

\subsection{Scenes}

The dataset comprises four distinct indoor public locations: ACS-EC, ACS-EG, IE-Central, and R-Central. Each scene presents unique challenges in terms of viewpoint, crowd density, and spatial layout, as illustrated in Figure~\ref{fig:samples}.

\textbf{ACS-EC} is a large multi-level atrium serving as a social hub. The camera captures a wide-angle view of the ground floor, featuring lounge seating, tables, and a commercial area, with an upper mezzanine level partially visible. This scene exhibits the highest crowd density in the dataset (${\approx}10-15$ persons per frame), many of whom are stationary or partially occluded by furniture. An additional elevated camera placement provides a top-down perspective, introducing viewpoint diversity.

\textbf{ACS-EG} is a long indoor corridor connecting building sections. The camera is positioned at ground level along the corridor axis, capturing $4-6$ persons per frame, typically walking toward or away from the camera. This scene is characterised by motion blur on nearby subjects and significant scale variation between near and far individuals.

\textbf{IE-Central} is the entrance hall of a separate building. The camera is mounted at an elevated angle, providing a broad top-down view of a tiled open floor with glass entry doors in the background. Crowd density is low to moderate ($7-10$ persons per frame), with people appearing at varying scales.

\textbf{R-Central} is a central atrium recorded from a high overhead angle. Prominent structural columns cause regular partial occlusion across the open floor space. Crowd density is consistently moderate and low-variance ($6-7$ persons per frame), making it the most uniform scene in the dataset.

\subsection{Frame Sampling and Video Splits}

The full dataset consists of $31$ videos distributed across the $4$ scenes. All videos were sampled at $5$\, fps, yielding $9,913$ frames in total. The choice of $5$\, fps was deliberate: on the one hand, it defines a low common denominator that ensures compatibility across different setups, since cameras and embedded sensors often operate at different rates; a relevant consideration for multi-sensor platforms such as robots, where sensor fusion pipelines must synchronise all input streams to a shared rate; on the other hand, it avoids redundant information by ensuring that consecutive frames are sufficiently distinct, as a person moves enough between samples to produce meaningfully different spatial configuration than near-duplicate appearances. Videos were recorded on different days to encourage variation in crowd density and lighting. The train/test split is performed at the video level, ensuring no temporal overlap between subsets, with each scene represented in both splits.

\subsection{Annotation Pipeline}

\textbf{Detection and segmentation subset.}
To efficiently scale our annotation across all $9,913$ frames while maintaining ground-truth fidelity, we employed a human-in-the-loop pipeline. Initial candidate masks and bounding boxes were generated using foundation models. Every frame then underwent human manual verification and correction. Annotators used SAM 2.1 to add missing masks, manually correcting imprecise mask boundaries via direct polygon labelling, and deleting false positives.

To quantitatively evaluate the quality of different foundation models for this initial candidate generation, we established a pure human-annotated control baseline. We isolated $20$ frames per video ($620$ frames total), prioritizing temporal diversity and avoiding near-duplicate frames. These control frames were annotated entirely from scratch by humans, without any pre-computed candidate priors. We then evaluated the 3 foundation models: SAM3~\cite{carion2026sam3}, GroundingSAM~\cite{ren2024groundedsamassemblingopenworld}, and EfficientGroundingSAM~\cite{xiong2023efficientsam} from the AutoDistill library~\cite{autodistil} against the $620$ frames human ground truth to benchmark their auto-labelling quality (detailed in Section~\ref{subsec:autolabel_quality}).

\textbf{Multi-object tracking subset.}
An additional $2{,}552$ frames were retained to form a MOT subset. Initial tracklets were generated from SAM3 detections, chosen for their high-recall coverage of all visible persons across time (Section~\ref{subsec:autolabel_quality}). Human reviewers then inspected every track, correcting identity switches, merging fragmented tracklets, removing ghost tracks, and linearly interpolating missing detections across short gaps. The subset follows the MOTChallenge format~\cite{dendorfer2021motchallenge} and is described in Section~\ref{subsec:mot_subset}.

\subsection{Ethics and Privacy}
Data collection was conducted under a formal institutional ethics approval letter, exclusively in publicly accessible campus areas. No audio was recorded, and no individuals were targeted. All faces were blurred prior to release using an automated de-identification pipeline; raw footage will not be released. Annotations encode only spatial and temporal information; no demographic attributes or personal identifiers are stored. The dataset will be released under a license restricting use to non-commercial computer vision research, explicitly prohibiting surveillance, re-identification of individuals from the images, or any application that could harm the individuals depicted.

\begin{table}[t]
\centering
\caption{Per-scene crowd statistics for the detection and segmentation subset. Density bins are defined as \textit{sparse} ($\leq3$), \textit{medium} (4--10), and \textit{dense} ($>10$) persons per frame. ACS-EC is a challenging scene, with 79.3\% of frames classified as dense and a mean of $12.23\pm3.80$ persons per frame. ACS-EG and R-Central are predominantly medium-density, while IE-Central spans the widest per-frame range (4--17 persons, 23.5\% dense frames). Occlusion rates are estimated via bounding-box overlap (IoU\,$>$\,0.1); ACS-EG shows the highest rate (38.3\%) despite its lower density, reflecting its corridor geometry and ground-level viewpoint rather than crowd size alone.}
\label{tab:scene_stats}
\resizebox{\columnwidth}{!}{%
\begin{tabular}{l r r r r r r}
\toprule
\multirow{2}{*}{\textbf{Scene}}
  & \multirow{2}{*}{\textbf{Instances}}
  & \multicolumn{1}{c}{\textbf{Persons / Frame}}
  & \multicolumn{3}{c}{\textbf{Density (\% frames)}}
  & \textbf{Occlusion} \\
\cmidrule(lr){3-3} \cmidrule(lr){4-6}
  & & Mean $\pm$ Std & Sparse & Medium & Dense & Rate \\
\midrule
ACS-EC     & 2{,}128 & $12.23 \pm 3.80$ &  2.9\% & 17.8\% & \textbf{79.3\%} & 27.5\% \\
ACS-EG     & 1{,}260 & $ 5.41 \pm 1.94$ & 18.5\% & \textbf{81.5\%} &  0.0\% & 38.3\% \\
IE-Central & 1{,}083 & $ 7.96 \pm 3.63$ &  0.0\% & 76.5\% & 23.5\% & 31.9\% \\
R-Central  &   391   & $ 6.86 \pm 1.52$ &  1.8\% & \textbf{98.2\%} &  0.0\% & 26.9\% \\
\midrule
\textbf{Overall} & \textbf{4{,}862} & $8.10 \pm 4.18$
               & 5.8\% & 62.0\% & 32.2\% & 30.3\% \\
\bottomrule
\end{tabular}%
}
\end{table}

\section{Dataset Statistics and Analysis}
\label{sec:dataset_stats}

\subsection{Crowd Density and Scene Diversity}
\label{subsec:crowd_density}
The $4$ scenes exhibit different crowd characteristics, showing diversity in the data. Most notably, ACS-EC is one of the challenging scenes with a mean of $12.2 \pm 3.8$ persons per frame and $79.3\%$ of frames classified as dense ($>$$10$ persons), which represents a highly congested indoor environment. While ACS-EG and R-Central are predominantly medium-density scenes, with zero dense frames and mean person counts of $5.4$ and $6.9$, respectively. IE-Central occupies an intermediate regime, with $23.5\%$ dense frames and the widest per-frame count range (4--17 persons) as shown in table~\ref{tab:scene_stats}. No scene contains empty frames; all footage captures active indoor occupancy.

\subsection{Scale and Aspect Ratio Variation}
\label{subsec:scale}

Table~\ref{tab:scale_stats} reports per-scene scale and aspect-ratio statistics revealing substantial variation driven by camera placement and room geometry. ACS-EG contains the largest instances (mean $135.6 \pm 78.2$\, px, $58.4$\ large) consistent with a corridor where subjects pass close to a fixed ground-level camera. On the opposite, ACS-EC and R-Central have small-to-medium instances (mean ${\approx}60$\, px), where detectors must localise persons at low resolution, known bottleneck for instance segmentation. IE-Central occupies an intermediate regime, with a near-even split between medium and large fractions (49.8\% and 46.9\%). Aspect ratios follow the same scene ordering, ranging from $2.02 \pm 0.89$ in ACS-EC to $3.28 \pm 1.03$ in ACS-EG. The elevated standard deviation in ACS-EC reflects the wider variety of poses, partial occlusions, and frame-boundary truncations characteristic of its dense atrium setting.

\begin{table*}[t]
\centering
\caption{Per-scene instance scale and aspect-ratio statistics. Relative scale is $\sqrt{A_{\text{box}}} / \sqrt{A_{\text{frame}}}$; absolute scale is the mean bounding-box side length in pixels. Size bins follow COCO convention: \textit{small} ${<}32^2$\,px, \textit{medium} $32^2$--$96^2$\,px, \textit{large} ${>}96^2$\,px. ACS-EG is dominated by large instances ($58.4$\%) due to its ground-level corridor viewpoint, while ACS-EC and R-Central present predominantly small-to-medium instances (mean ${\approx}60$\,px), posing a challenge for instance segmentation. The high aspect-ratio variance in ACS-EC ($2.02 \pm 0.89$) reflects diverse poses, partial occlusions, and frame-boundary truncations in its dense atrium setting.}
\label{tab:scale_stats}
\resizebox{\linewidth}{!}{%
\begin{tabular}{l r r r r r r}
\toprule
\multirow{2}{*}{\textbf{Scene}}
  & \multicolumn{2}{c}{\textbf{Relative Scale}}
  & \textbf{Abs.\ Scale (px)}
  & \multicolumn{3}{c}{\textbf{Size Distribution (\% instances)}} \\
\cmidrule(lr){2-3} \cmidrule(lr){5-7}
  & Mean $\pm$ Std & AR Mean $\pm$ Std & Mean $\pm$ Std & Small & Medium & Large \\
\midrule
ACS-EC     & $0.063 \pm 0.039$ & $2.02 \pm 0.89$ & $ 60.8 \pm  37.7$ & 17.2\% & 73.5\% &  9.4\% \\
ACS-EG     & $0.141 \pm 0.081$ & $3.28 \pm 1.03$ & $135.6 \pm  78.2$ &  1.0\% & 40.6\% & \textbf{58.4\%} \\
IE-Central & $0.095 \pm 0.040$ & $2.48 \pm 0.66$ & $ 91.2 \pm  38.6$ &  3.3\% & 49.8\% & 46.9\% \\
R-Central  & $0.062 \pm 0.017$ & $2.26 \pm 0.73$ & $ 59.1 \pm  16.4$ &  4.9\% & \textbf{93.6\%} &  1.5\% \\
\bottomrule
\end{tabular}%
}
\end{table*}

\begin{figure*}[t]
    \centering
    \begin{subfigure}[t]{0.24\textwidth}
        \includegraphics[width=\textwidth]{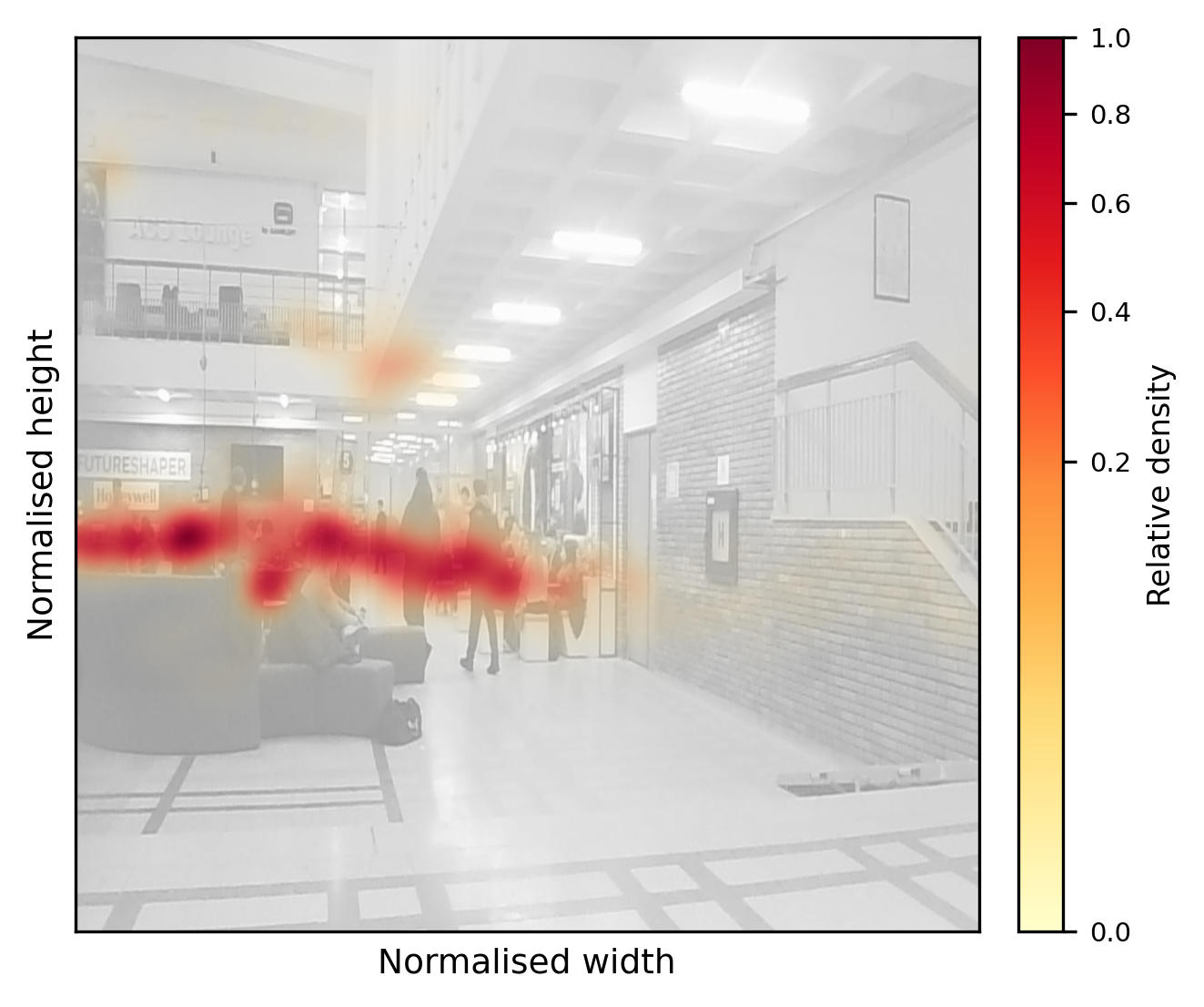}
        \caption{ACS-EC: dense atrium with mixed mobile and stationary occupants; the highest crowd density (79.3\% dense frames) in the dataset.}
        \label{fig:heatmap_acs_ec}
    \end{subfigure}
    \hfill
    \begin{subfigure}[t]{0.24\textwidth}
        \includegraphics[width=\textwidth]{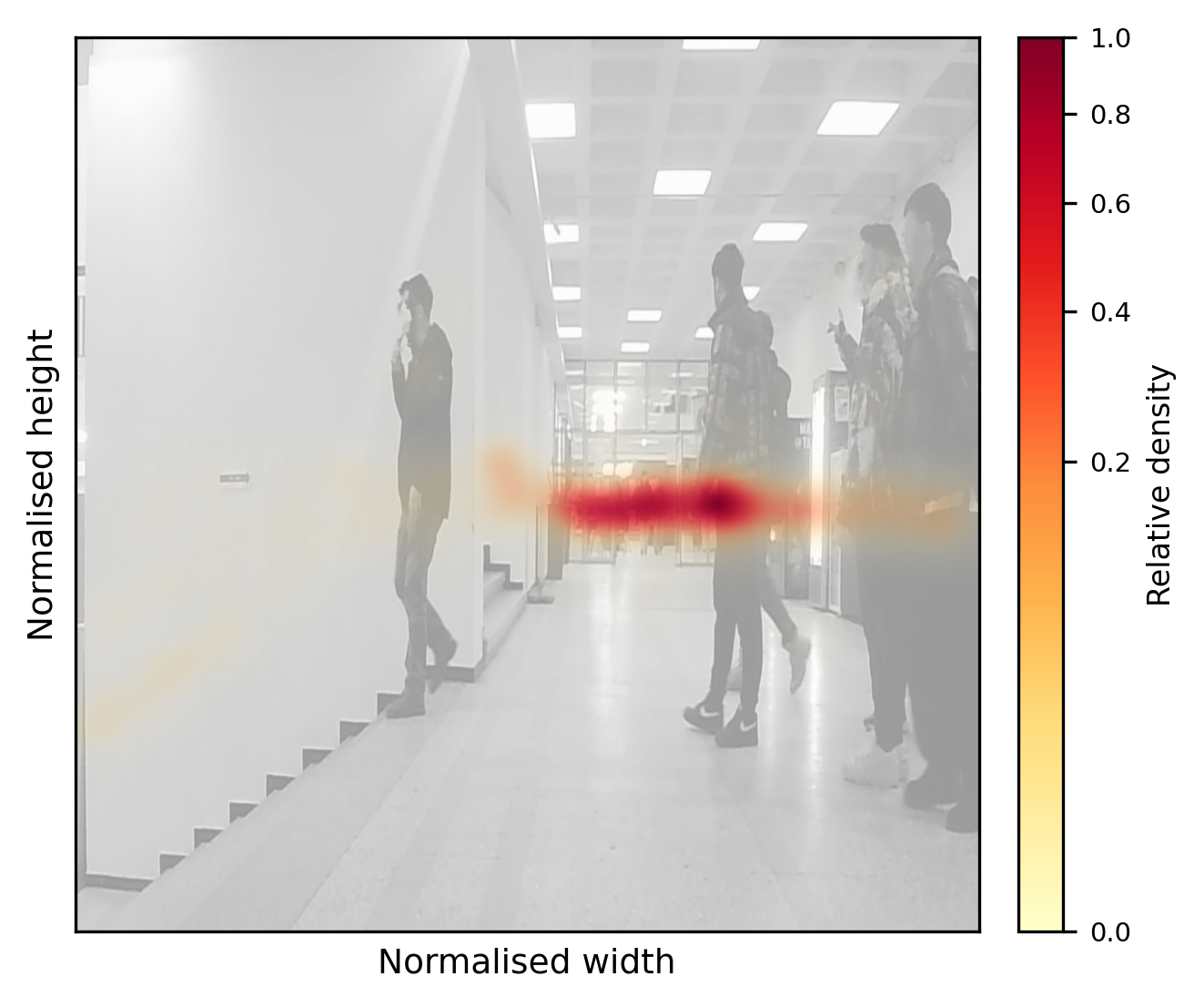}
        \caption{ACS-EG: narrow corridor with mid-depth density cluster and strong near-to-distal scale variance.}
        \label{fig:heatmap_acs_eg}
    \end{subfigure}
    \hfill
    \begin{subfigure}[t]{0.24\textwidth}
        \includegraphics[width=\textwidth]{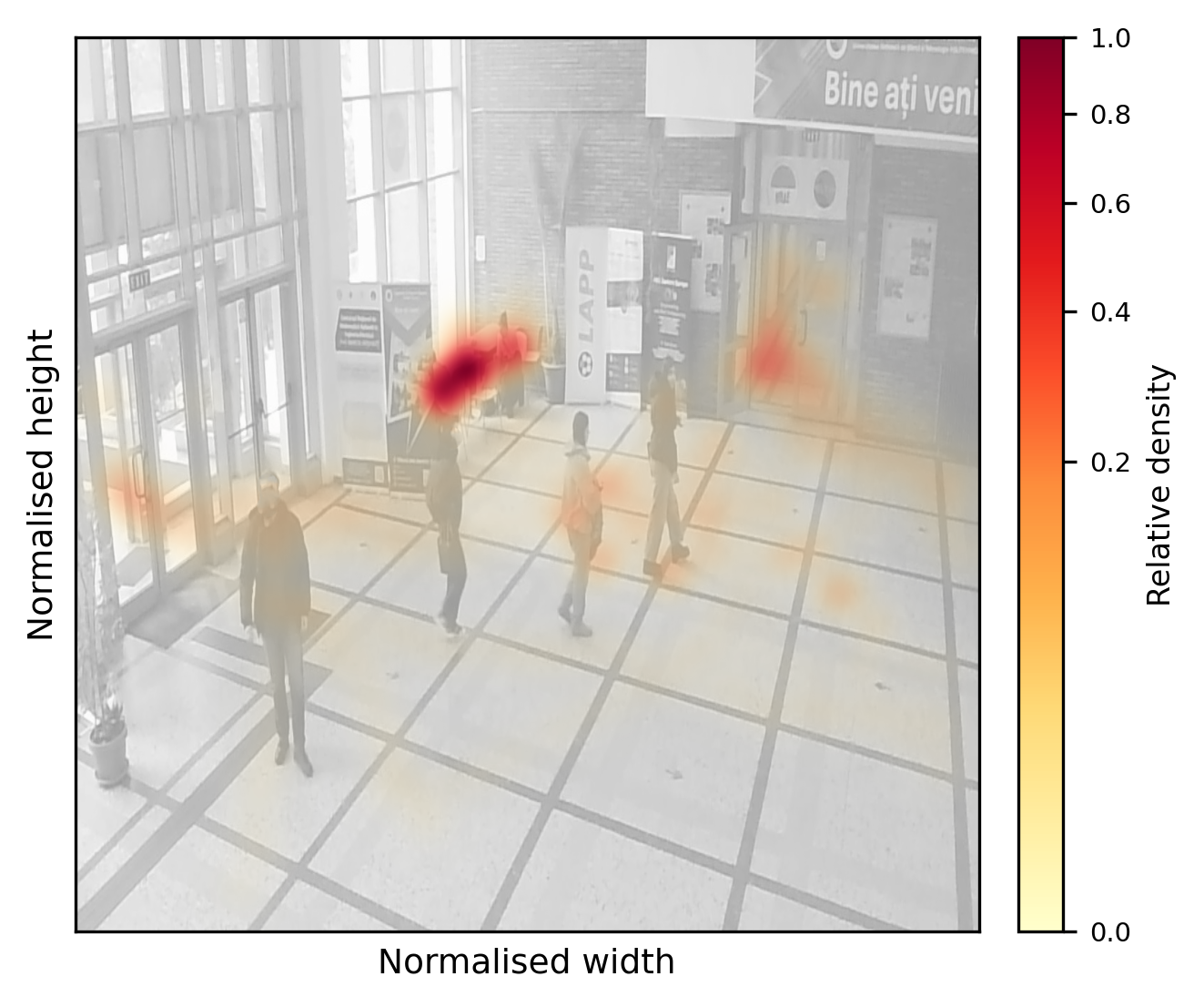}
        \caption{IE-Central: entrance hall with density split across entry, corridor junction, and seating zones (4--17 persons/frame).}
        \label{fig:heatmap_ie_central}
    \end{subfigure}
    \hfill
    \begin{subfigure}[t]{0.24\textwidth}
        \includegraphics[width=\textwidth]{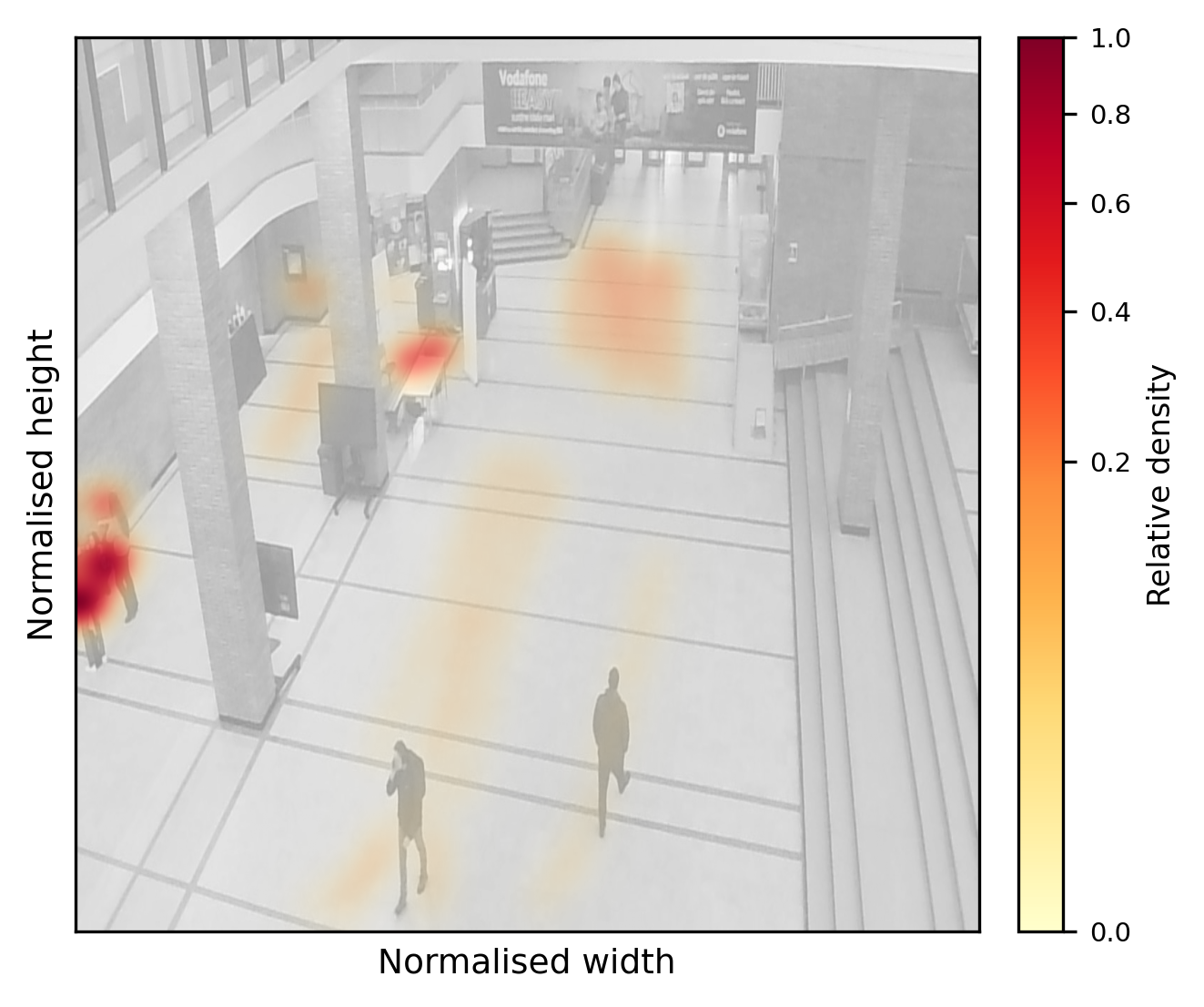}
        \caption{R-Central: overhead view of column-interrupted atrium with diffuse density and ambiguous vertical pedestrian flow.}
        \label{fig:heatmap_r_central}
    \end{subfigure}

    \caption{Spatial density heatmaps showing the normalised distribution of person bounding-box centres across all annotated frames per scene. Colour encodes relative density (yellow $\to$ dark red = low $\to$ high). The heatmaps reveal four distinct spatial regimes: a dominant horizontal band in ACS-EC driven by circulation traffic and stationary occupants in the common area; a concentrated mid-depth cluster in ACS-EG reflecting strong scale variance along its linear corridor axis; three discrete zones in IE-Central spanning the entry, corridor junction, and seating area; and a diffuse, column-interrupted spread in R-Central where the overhead viewpoint collapses ascending and descending pedestrian flow into a single projection.  These spatial patterns directly inform the per-scene variation in crowd density, occlusion rate, and detection difficulty reported in  Sections~\ref{subsec:occlusion} and~\ref{subsec:autolabel_quality}.}
    \label{fig:density_heatmaps}
\end{figure*}

\subsection{Occlusion Analysis}
\label{subsec:occlusion}

Occlusion rates were estimated using a bounding-box overlap proxy: an instance is flagged as occluded if its box overlaps with at least one other annotation in the same frame at IoU~$>0.1$. It is worth noting that this proxy exclusively considers overlaps between annotated pedestrian instances; occlusions caused by static environmental elements such as pillars, sofas, or other scene furniture are not captured by this metric, meaning that the reported rates likely underestimate the true level of occlusion present in the scene. Overall, $30.3\%$ of all annotated instances are subject to occlusion. Notably, ACS-EG shows the highest occlusion rate ($38.3\%$) despite being the least densely populated scene, indicating that occlusion is governed not only by crowd density but also by camera angle, distance and corridor geometry. ACS-EC and R-Central share similar rates (${\approx}27\%$), while IE-Central falls between the two at $31.9\%$.

\subsection{Auto-Labelling Quality}
\label{subsec:autolabel_quality}

\begin{figure*}[t]
    \centering
    \includegraphics[width=\linewidth]{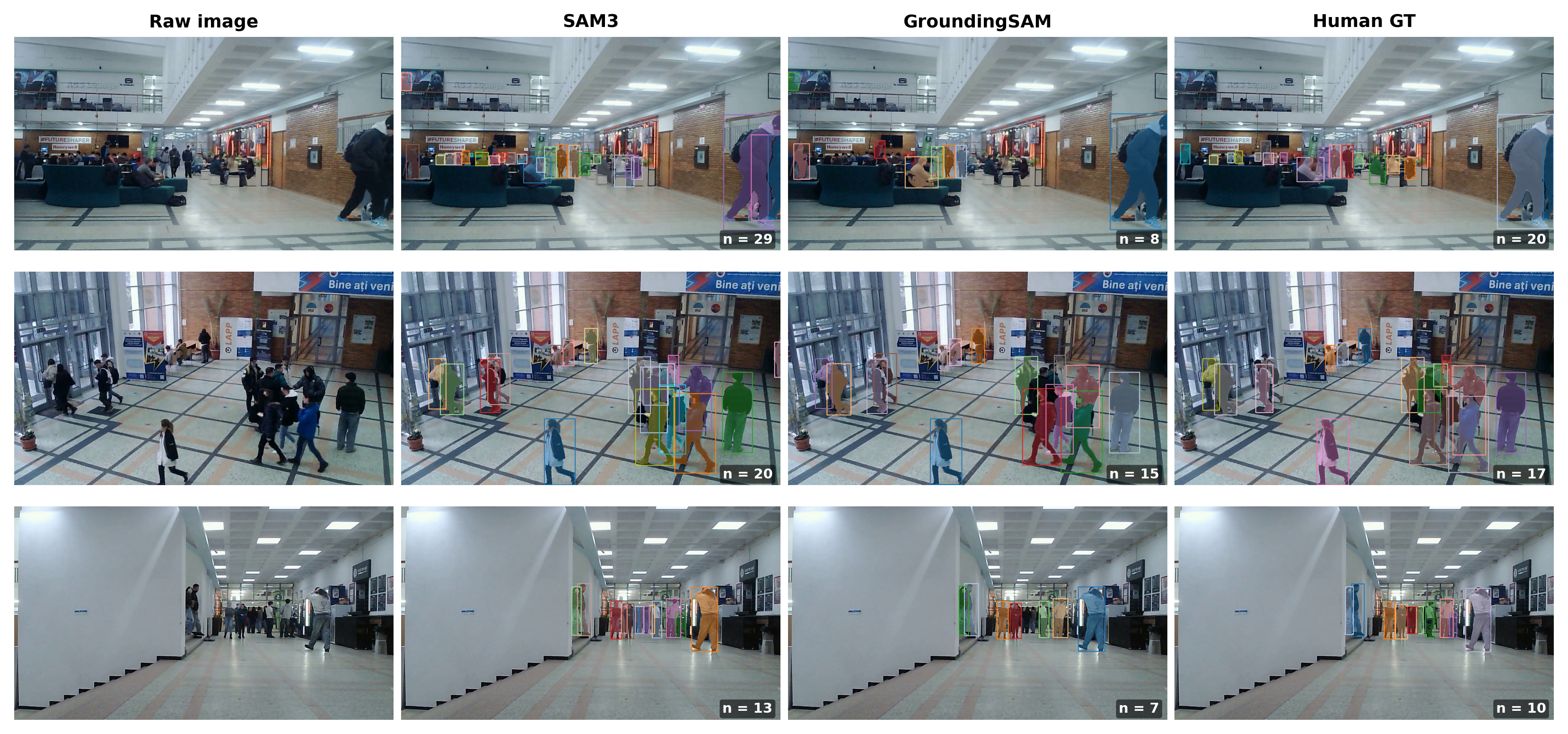}
    \caption{Qualitative comparison of auto-labelling methods across ACS-EC, IE-Central, and R-Central (rows, top to bottom). Columns show the raw image, SAM3, GroundingSAM, and human ground truth, with per-frame instance counts (n). SAM3 produces false positives on ACS-EC (row 1); GroundingSAM misses occluded persons across all scenes.}
    \label{fig:scenes}
\end{figure*}

We evaluated the $3$ automatic annotation methods against the human ground truth on all $620$ labelled frames. Results are reported in Table~\ref{tab:autolabel_results}. SAM3 achieves the highest recall across all scenes ($0.88-0.98$ at IoU\,0.5), but over-predicts: ACS-EC generates $3,596$ predictions against $2,128$ ground-truth instances, yielding a precision of only $0.52$. This high-recall, low-precision profile makes SAM3 the optimal starting point for human correction, minimising missed persons at the cost of easily removable false positives. GroundingSAM and EfficientGroundingSAM behave near-identically across all metrics and scenes, showing that the efficient variant preserves annotation quality; both are considerably more conservative than SAM3, achieving higher precision ($0.83$--$0.93$) at the cost of lower recall ($0.47$--$0.94$ depending on scene).

\begin{table*}[t]
\centering
\caption{Auto-labelling quality against human ground truth across all four scenes. $N_{\text{pred}}$ and $N_{\text{GT}}$ denote predicted and ground-truth instance counts; Cohen's $\kappa$ follows \cite{landis1977application}. SAM3 maximises recall at the cost of precision; GroundingSAM and EfficientGroundingSAM behave near-identically with the inverse trade-off. All methods degrade most on ACS-EC.}
\label{tab:autolabel_results}
\resizebox{\textwidth}{!}{%
\begin{tabular}{ll rr rr rr r r}
\toprule
\multirow{2}{*}{\textbf{Scene}}
  & \multirow{2}{*}{\textbf{Method}}
  & \multirow{2}{*}{$N_{\text{pred}}$}
  & \multirow{2}{*}{$N_{\text{GT}}$}
  & \multicolumn{2}{c}{\textbf{AP} $\uparrow$}
  & \multicolumn{2}{c}{\textbf{Prec. $\uparrow$ / Rec. $\uparrow$ @\,0.5}}
  & \textbf{Mask IoU} $\uparrow$
  & \textbf{Cohen's} $\uparrow$ \\
\cmidrule(lr){5-6} \cmidrule(lr){7-8}
  & & & & @0.5 & @0.75 & P & R & (mean) & $\kappa$ \\
\midrule
\multirow{3}{*}{\textit{ACS-EC}}
  & SAM3   & 3596 & \multirow{3}{*}{2128} & \textbf{0.783} & \textbf{0.599} & 0.522          & \textbf{0.882} & 0.802          & \textbf{0.849} \\
  & G-SAM  & 1214 &                       & 0.435          & 0.324          & \textbf{0.831} & 0.474          & \textbf{0.831} & 0.762          \\
  & EG-SAM & 1212 &                       & 0.433          & 0.323          & 0.829          & 0.472          & 0.803          & 0.758          \\
\addlinespace[2pt]
\multirow{3}{*}{\textit{ACS-EG}}
  & SAM3   & 1643 & \multirow{3}{*}{1260} & \textbf{0.963} & \textbf{0.862} & 0.749          & \textbf{0.976} & 0.859          & 0.926          \\
  & G-SAM  & 1359 &                       & 0.922          & 0.824          & 0.870          & 0.939          & \textbf{0.869} & 0.931          \\
  & EG-SAM & 1360 &                       & 0.924          & 0.826          & \textbf{0.872} & 0.941          & 0.850          & \textbf{0.934} \\
\addlinespace[2pt]
\multirow{3}{*}{\textit{IE-Central}}
  & SAM3   & 1235 & \multirow{3}{*}{1083} & \textbf{0.928} & \textbf{0.788} & 0.829          & \textbf{0.946} & \textbf{0.836} & \textbf{0.923} \\
  & G-SAM  &  953 &                       & 0.803          & 0.670          & \textbf{0.929} & 0.817          & 0.835          & 0.902          \\
  & EG-SAM &  954 &                       & 0.804          & 0.672          & \textbf{0.929} & 0.818          & 0.809          & 0.904          \\
\addlinespace[2pt]
\multirow{3}{*}{\textit{R-Central}}
  & SAM3   & 589  & \multirow{3}{*}{391}  & \textbf{0.946} & \textbf{0.832} & 0.652          & \textbf{0.982} & 0.847          & \textbf{0.864} \\
  & G-SAM  & 551  &                       & 0.901          & 0.774          & \textbf{0.670} & 0.944          & \textbf{0.853} & 0.859          \\
  & EG-SAM & 550  &                       & 0.900          & 0.774          & 0.669          & 0.941          & 0.830          & 0.849          \\
\bottomrule
\end{tabular}%
}
\end{table*}

All three methods drop sharply on ACS-EC: AP@0.5 falls from $0.90$--$0.96$ on other scenes to $0.43$--$0.78$, consistent with its denser frames ($79.3$\%), smaller instances (mean $60.8$,px), and higher occlusion ($27.5$\%). Cohen’s $\kappa$ shows the same pattern ($0.76$--$0.85$ on ACS-EC vs. $0.90$--$0.93$ elsewhere): all results indicate strong agreement ($\kappa>0.80$) except GroundingSAM and EfficientGroundingSAM on ACS-EC ($0.76$--$0.78$), which lie on the moderate–strong boundary \cite{landis1977application}. Mask IoU stays high ($0.80$--$0.87$), suggesting corrections mainly address missed detections and false positives rather than mask refinement.

\begin{table*}[htpb]
\centering
\caption{Detection and segmentation benchmark results. RT-DETR-L achieves the highest accuracy, but at a high compute cost; YOLOv8n-seg offers the best accuracy--efficiency trade-off for the combined detection and segmentation task. Latency measured on an NVIDIA RTX\,4060Ti (16\,GB, batch size 1).}

\label{tab:model_results}
\resizebox{\textwidth}{!}{%
\begin{tabular}{ll rrrr r rrr rrr}
\toprule
\multirow{2}{*}{\textbf{Model}}
  & \multirow{2}{*}{\textbf{Task}}
  & \multicolumn{2}{c}{\textbf{Box mAP} $\uparrow$}
  & \multicolumn{2}{c}{\textbf{Mask mAP} $\uparrow$}
  & \textbf{Mean}
  & \multicolumn{3}{c}{\textbf{Detection} $\uparrow$}
  & \textbf{Latency}
  & \textbf{Size}
  & \textbf{GFLOPs} \\
\cmidrule(lr){3-4} \cmidrule(lr){5-6} \cmidrule(lr){8-10}
  & & @0.5 & @0.50:95 & @0.5 & @0.50:95 & IoU & P & R & F1 & (ms) & (MB) & \\
\midrule
YOLOv8n        & detect  & 0.864 & 0.616          & ---            & ---            & ---            & 0.876          & 0.766          & 0.817          & 2.28           & 6.22           & 8.1            \\
YOLOv26n       & detect  & 0.796          & 0.560          & ---            & ---            & ---            & 0.749          & 0.715          & 0.732          & 2.81           & \textbf{5.36}  & \textbf{5.2}   \\
RT-DETR-L      & detect  & \textbf{0.911} & \textbf{0.704} & ---            & ---            & ---            & 0.898          & \textbf{0.834} & \textbf{0.865} & 27.36          & 66.21          & 103.4          \\
YOLOv8n-seg    & segment & 0.864          & 0.620          & \textbf{0.833} & \textbf{0.541} & \textbf{0.630} & \textbf{0.915} & 0.757          & 0.829 & \textbf{1.89}  & 6.76           & 11.3           \\
YOLOv26n-seg   & segment & 0.808          & 0.572          & 0.787          & 0.512          & 0.599          & 0.773          & 0.741          & 0.757          & 2.61           & 6.51           & 9.0            \\
\bottomrule
\end{tabular}%
}
\end{table*}

\subsection{MOT Subset}
\label{subsec:mot_subset}

\begin{table}[htpb]
\centering
\caption{Per-scene track statistics comparing human-annotated ground truth  against SAM3-Native (raw SAM3 tracklets) and SAM3-BotSort (SAM3 detections  linked by BoT-SORT). \textbf{Mean}, \textbf{Med.}, \textbf{Min}, and  \textbf{Max} refers to the track length in frames. Human tracks are consistently fewer and longer than automatic counterparts, reflecting the consolidation of fragmented and spurious tracklets during human review.}
\label{tab:tracking_stats}
\resizebox{\columnwidth}{!}{%
\setlength{\tabcolsep}{4pt}
\begin{tabular}{l r r r r r r}
\toprule
\textbf{Method} & \textbf{Uniq.\ IDs} & \textbf{Tracks} & \textbf{Mean} & \textbf{Med.} & \textbf{Min} & \textbf{Max} \\
\midrule
\multicolumn{7}{l}{\textit{ACS-EC}} \\
\midrule
Human        & \textbf{260} & \textbf{260} & 32.51          & 31.0          & \textbf{1} & \textbf{102} \\
SAM3-Native  & 321          & 321          & \textbf{40.90} & \textbf{39.0} & \textbf{1} & 82           \\
SAM3-BotSort & 507          & 507          & 24.36          & 17.0          & \textbf{1} & 82           \\
\midrule
\multicolumn{7}{l}{\textit{ACS-EG}} \\
\midrule
Human        & \textbf{122} & \textbf{122} & \textbf{33.49} & \textbf{26.5} & \textbf{1} & \textbf{136} \\
SAM3-Native  & 193          & 193          & 29.22          & 22.0          & \textbf{1} & \textbf{136} \\
SAM3-BotSort & 294          & 294          & 17.36          &  8.0          & \textbf{1} & \textbf{136} \\
\midrule
\multicolumn{7}{l}{\textit{IE-Central}} \\
\midrule
Human        & \textbf{77}  & \textbf{77}  & 56.92          & 53.0          & \textbf{1} & \textbf{379} \\
SAM3-Native  & 104          & 104          & \textbf{59.80} & \textbf{45.0} & \textbf{1} & 333          \\
SAM3-BotSort & 186          & 186          & 31.19          & 14.5          & \textbf{1} & \textbf{379} \\
\midrule
\multicolumn{7}{l}{\textit{R-Central}} \\
\midrule
Human        & \textbf{33}  & \textbf{33}  & 77.85          & 76.0          & 5           & \textbf{156} \\
SAM3-Native  & 41           & 41           & \textbf{99.17} & \textbf{93.0} & \textbf{13} & \textbf{156} \\
SAM3-BotSort & 64           & 64           & 62.62          & 47.5          & \textbf{1}  & \textbf{156} \\
\bottomrule
\end{tabular}%
}
\end{table}
\vspace{-4pt}

Beyond static detection and segmentation, we provide a multi-object tracking subset comprising $2,552$ frames derived from the same $31$ videos. Table~\ref{tab:tracking_stats} compares the track statistics of the human-annotated ground truth against $2$ automatic sources: SAM3-Native~\cite{carion2026sam3} (raw SAM3 tracklets) and SAM3-BotSort~\cite{aharon2022bot} (SAM3 detections associated by BoT-SORT). Human-annotated tracks are consistently fewer and longer than their automatic counterparts across all scenes, reflecting the removal of fragmented and spurious tracks during the review stage. Table~\ref{tab:tracking_postproc} quantifies the corrections applied during human review. The $2$ automatic sources present complementary failure modes: SAM3-BotSort required substantially more intervention, most notably in ACS-EC, where $164$ ghost tracks were deleted, and $118$ merges were performed, yet its longer initial tracklets (mean $24.4$ frames vs.\ $40.9$ for SAM3-Native in ACS-EC) confirm that BoT-SORT produces more temporally coherent associations that are easier to correct. SAM3-Native, by contrast, generates shorter and noisier tracks that require more frame-level interpolation (e.g.\ $1,067$ frames in ACS-EC vs.\ $596$ for SAM3-BotSort), making it a more labour-intensive starting point despite requiring fewer structural edits.

\begin{table}[htbp]
\centering
\caption{Per-scene human correction actions applied to SAM3-Native and  SAM3-BotSort tracklets during human review. $\Delta$IDs and $\Delta$Trk are the net reductions in unique identities and track count after review. $\Delta$Len is the mean change in track length (frames); negative values indicate that human tracks are shorter on average than the automatic source. \textit{Ghosts} are fully deleted spurious tracks; \textit{Merged} and  \textit{ID Sw.} count identity-switch corrections; \textit{Interp.\ Frames}  is the number of missing detections filled by linear interpolation. SAM3-BotSort consistently required more structural edits ($\Delta$IDs, Ghosts), while SAM3-Native demanded heavier interpolation, reflecting their complementary failure modes.}
\label{tab:tracking_postproc}
\resizebox{\columnwidth}{!}{%
\setlength{\tabcolsep}{4pt}
\begin{tabular}{l r r r r r r r}
\toprule
 & \multicolumn{3}{c}{\textbf{Track Changes}} & \multicolumn{3}{c}{\textbf{Refinement Actions}} & \textbf{Interp.} \\
\cmidrule(lr){2-4} \cmidrule(lr){5-7}
\textbf{Method} & $\Delta$IDs & $\Delta$Trk & $\Delta$Len & Ghosts & Merged & ID Sw. & Frames \\
\midrule
\multicolumn{8}{l}{\textit{ACS-EC}} \\
\midrule
SAM3-Native  &  61          &  61          & $-$8.39        &  41          & \textbf{85}  & \textbf{85}  & \textbf{1067} \\
SAM3-BotSort & \textbf{247} & \textbf{247} & \textbf{8.15}  & \textbf{164} & 118          & 118          &  596          \\
\midrule
\multicolumn{8}{l}{\textit{ACS-EG}} \\
\midrule
SAM3-Native  &  71          &  71          &  4.27          &  36          & \textbf{49}  & \textbf{49}  & \textbf{336}  \\
SAM3-BotSort & \textbf{172} & \textbf{172} & \textbf{16.13} & \textbf{105} & 72           & 72           &  93           \\
\midrule
\multicolumn{8}{l}{\textit{IE-Central}} \\
\midrule
SAM3-Native  &  27          &  27          & $-$2.88        &  20          & \textbf{27}  & \textbf{27}  & \textbf{1010} \\
SAM3-BotSort & \textbf{109} & \textbf{109} & \textbf{25.73} & \textbf{90}  & 25           & 25           &  134          \\
\midrule
\multicolumn{8}{l}{\textit{R-Central}} \\
\midrule
SAM3-Native  &   8          &   8          & $-$21.32       &   5          & \textbf{11}  & \textbf{11}  & \textbf{301}  \\
SAM3-BotSort & \textbf{31}  & \textbf{31}  & \textbf{15.22} & \textbf{21}  & 10           & 10           &   0           \\
\bottomrule
\end{tabular}%
}
\end{table}

\section{Benchmarks}
\label{sec:benchmarks}

\subsection{Object Detection and Segmentation}
Table~\ref{tab:model_results} reports detection and segmentation performance for $5$ model configurations. RT-DETR-L~\cite{zhao2024detrs} achieves the highest box mAP@0.5 ($0.911$) and mAP@0.50:0.95 ($0.704$) but at a compute cost ($27.36$\,ms latency, $103.4$\,GFLOPs). Among YOLO variants, YOLOv8n-seg is the strongest all-round choice, matching YOLOv8n~\cite{sohan2024review} detection accuracy while adding instance masks (mask mAP@0.5\,=\,$0.833$) at the lowest latency of any model ($1.89$\, ms); YOLOv26n~\cite{sapkota2025yolo26} offers the smallest footprint ($5.36$\, MB, $5.2$\, GFLOPs) at a moderate accuracy cost, suited to resource-constrained deployment. The RT-DETR-L advantage is most pronounced on ACS-EC, where small-scale, densely packed instances disproportionately challenge lightweight detectors, show it as the benchmark's challenging scene (Section~\ref{subsec:scale}).

All models were trained on 2 of the 4 scenes (ACS-EC and ACS-EG), while evaluation was performed on the held-out scenes. To ensure a fair comparison across architectures, we used a single training pipeline and kept optimization and augmentation settings fixed for all runs. Models were initialized from COCO-pretrained weights~\cite{sapkota2025ultralytics} and trained for 30 epochs with batch size 16 at an input resolution $640\times640$. The learning rate followed a linear schedule with a base learning rate of $lr=0.01$ and final factor $lrf=0.01$, with a 3-epoch warmup; the optimizer was left to the Ultralytics~\cite{sapkota2025ultralytics} default selection. Data augmentation was identical across all runs and included HSV jitter, geometric transforms (translation = 0.1, scale = 0.5, horizontal flip probability = 0.5) and mosaic augmentation, closed in the last 10 epochs. The dataset was converted from COCO-style to YOLO format~\cite{vijayakumar2024yolo} and trained as a single-class problem.

\subsection{Multi-Object Tracking}
We benchmark six detector--tracker combinations (YOLOv8n and RT-DETR-L paired with ByteTrack, BoT-SORT, and OC-SORT) across all four scenes; results are reported in Table~\ref{tab:mot_results}.

\begin{table}[htpb]
\centering
\caption{Multi-object tracking results per scene. Best value per metric
per scene is \textbf{bolded}. MT\%: mostly tracked; ML\%: mostly lost;
IDS: identity switches.}
\label{tab:mot_results}
\resizebox{\columnwidth}{!}{%
\setlength{\tabcolsep}{4pt}
\begin{tabular}{ll l rr rr r r}
\toprule
\textbf{Scene} & \textbf{Detector} & \textbf{Tracker}
  & \textbf{MOTA}$\uparrow$ & \textbf{IDF1}$\uparrow$
  & \textbf{MT\%}$\uparrow$ & \textbf{ML\%}$\downarrow$
  & \textbf{IDS}$\downarrow$ & \textbf{FPS}$\uparrow$ \\
\midrule
\multirow{6}{*}{\textit{Overall}}
  & YOLOv8n   & ByteTrack & 48.5 & 66.9 & 49.0 & 35.4 & 161          & \textbf{108.2} \\
  & YOLOv8n   & BotSort   & 51.5 & 68.6 & 52.9 & 34.6 & 143          & 118.5          \\
  & YOLOv8n   & OC-SORT   & 51.4 & 68.6 & 45.3 & 35.8 & \textbf{138} & 119.5          \\
  & RT-DETR-L & ByteTrack & 51.4 & 69.3 & 55.5 & 27.2 & 172          & 31.4           \\
  & RT-DETR-L & BotSort   & 55.5 & 71.6 & \textbf{60.6} & \textbf{26.2} & 161 & 32.0  \\
  & RT-DETR-L & OC-SORT   & \textbf{56.2} & \textbf{71.8} & 54.7 & 27.4 & 166 & 32.7  \\
\midrule
\multirow{6}{*}{\textit{ACS-EC}}
  & YOLOv8n   & ByteTrack & 33.8 & 57.8 & 37.3 & 51.2 & 57           & 94.8  \\
  & YOLOv8n   & BotSort   & 35.1 & 58.3 & 38.1 & 51.5 & \textbf{49}  & \textbf{107.4}          \\
  & YOLOv8n   & OC-SORT   & 35.3 & 57.3 & 34.2 & 51.9 & 56           & 104.4          \\
  & RT-DETR-L & ByteTrack & 39.1 & 62.5 & \textbf{45.0} & 40.0 & 59  & 30.6           \\
  & RT-DETR-L & BotSort   & \textbf{40.2} & \textbf{63.3} & \textbf{45.0} & \textbf{38.9} & 53 & 31.1 \\
  & RT-DETR-L & OC-SORT   & 40.0 & 61.3 & 41.5 & 38.9 & 66           & 31.7           \\
\midrule
\multirow{6}{*}{\textit{ACS-EG}}
  & YOLOv8n   & ByteTrack & 75.0 & 78.7 & 68.9 & 18.9 & 62           & 109.7          \\
  & YOLOv8n   & BotSort   & 79.0 & 81.1 & 75.4 & 17.2 & 54           & 124.3          \\
  & YOLOv8n   & OC-SORT   & 75.8 & 82.8 & 59.0 & 19.7 & \textbf{40}  & \textbf{124.5} \\
  & RT-DETR-L & ByteTrack & 75.3 & 80.7 & 71.3 & 15.6 & 52           & 32.0           \\
  & RT-DETR-L & BotSort   & \textbf{79.4} & 82.9 & \textbf{81.2} & \textbf{13.1} & 52 & 32.1 \\
  & RT-DETR-L & OC-SORT   & 77.9 & \textbf{84.0} & 70.5 & 15.6 & 44  & 33.1           \\
\midrule
\multirow{6}{*}{\textit{IE-Central}}
  & YOLOv8n   & ByteTrack & 73.4 & 79.3 & 58.4 & 14.3 & 28           & 120.1 \\
  & YOLOv8n   & BotSort   & 78.6 & 82.6 & 68.8 & 11.7 & 28           & 118.6          \\
  & YOLOv8n   & OC-SORT   & 77.3 & 82.5 & 63.6 & 13.0 & \textbf{19}  & \textbf{121.0}          \\
  & RT-DETR-L & ByteTrack & 71.2 & 77.6 & 62.3 & 10.4 & 35           & 32.0           \\
  & RT-DETR-L & BotSort   & \textbf{81.4} & 83.2 & \textbf{76.6} & \textbf{9.1} & 36 & 32.5 \\
  & RT-DETR-L & OC-SORT   & 79.9 & \textbf{84.0} & 72.7 & 11.7 & 32  & 33.4           \\
\midrule
\multirow{6}{*}{\textit{R-Central}}
  & YOLOv8n   & ByteTrack & 12.4 & 56.8 & 45.5 & 21.2 & 14           & 108.3          \\
  & YOLOv8n   & BotSort   & 15.3 & 58.1 & 48.5 & 18.2 & \textbf{12}  & 117.4          \\
  & YOLOv8n   & OC-SORT   & 20.9 & 58.1 & 39.4 & 21.2 & 23           & \textbf{130.0} \\
  & RT-DETR-L & ByteTrack & 20.5 & 60.4 & 63.6 & \textbf{9.1} & 26   & 28.5           \\
  & RT-DETR-L & BotSort   & 23.2 & 61.9 & \textbf{69.7} & \textbf{9.1} & 20 & 32.7   \\
  & RT-DETR-L & OC-SORT   & \textbf{34.0} & \textbf{65.0} & 57.6 & 18.2 & 24 & 31.6  \\
\bottomrule
\end{tabular}%
}
\end{table}

\textbf{Detector impact.} RT-DETR-L consistently outperforms YOLOv8n across all tracker and scene combinations. The gain is mostly notable on ACS-EC, where MOTA improves from $35.3$ to $40.2$ with BoT-SORT, confirming that detection quality is the primary bottleneck on this dataset rather than the association algorithm.

\textbf{Tracker comparison. }RT-DETR-L\,+\,OC-SORT~\cite{cao2023observation} achieves the best overall MOTA ($56.2$), while BoT-SORT consistently yields the best IDF1 and the lowest identity-switch count across both detectors, indicating better identity preservation. For latency-constrained deployment, YOLOv8n\,+\,ByteTrack~\cite{zhang2022bytetrack} exceeds $108$\,FPS while remaining competitive on MOTA.

\textbf{Scene difficulty. }ACS-EC is a challenging tracking scene, with MOTA peaking at $40.2$ even under RT-DETR-L, driven by high density, small scale, and frequent occlusion. R-Central has a hidden complexity: despite moderate density, the overhead viewpoint and structural columns produce abrupt appearance changes that disproportionately elevate identity switches. ACS-EG and IE-Central achieve MOTA above $0.70$ across all detector--tracker configurations, reflecting their more favourable density and scale conditions.

\section{Conclusion}
\label{sec:conclusion}
To address the need for human identification, we introduce a new indoor dataset for human detection, instance segmentation, and multi-object tracking, capturing $31$ videos across four scenes with diverse crowd density, camera geometry, and occlusion patterns. The dataset provides $620$ manually annotated frames with per-instance masks and a $2,552$-frame tracking subset with human-verified identities. Our systematic auto-labelling quality study applies Cohen's $\kappa$, AP, and mask IoU across multiple foundation models in a crowded indoor setting—shows that SAM3 is the optimal starting point for human correction due to its high recall, while EfficientGroundingSAM achieves comparable quality to GroundingSAM at lower inference cost. Detection and tracking baselines confirm that ACS-EC is the hardest scene due to high density, small scale, and occlusion, and that RT-DETR-L + OC-SORT provides the best tracking accuracy while YOLOv8n + ByteTrack offers a strong real-time alternative. Limitations include the modest size of the human-annotated subset ($620$ frames) and the single-institution data source, which may limit generalisation across building types. Future work will expand to additional indoor environments, explore night and low-light conditions, and extend annotations to support person re-identification. The dataset, annotations, and annotation pipeline scripts are publicly available at \url{https://sheepseb.github.io/IndoorCrowd/}. 

\section*{Acknowledgments}
This work was supported by the project “Romanian Hub for Artificial Intelligence - HRIA”, Smart Growth, Digitization and Financial Instruments Program, 2021-2027, MySMIS no. 351416

{
    \small
    \bibliographystyle{ieeenat_fullname}
    \bibliography{main}

@String(CVPR= {IEEE Conf. Comput. Vis. Pattern Recog.})

@String(ICCV= {Int. Conf. Comput. Vis.})

@String(ECCV= {Eur. Conf. Comput. Vis.})

@String(TMM  = {IEEE Trans. Multimedia})

@String(CVPR  = {CVPR})

@String(ICCV  = {ICCV})

@String(ECCV  = {ECCV})

@String(TMM   =	{IEEE TMM})

@article{carion2026sam3,
  title={SAM 3: Segment Anything with Concepts},
  author={Carion, Nicolas and Gustafson, Laura and Hu, Yuan-Ting and Debnath, Shoubhik and Hu, Ronghang and others},
  journal={arXiv preprint arXiv:2511.16719},
  year={2026}
}

@inproceedings{kirillov2023sam,
  title={Segment Anything},
  author={Kirillov, Alexander and Mintun, Eric and Ravi, Nikhila and Mao, Hanzi and Rolland, Chloe and others},
  booktitle={Proceedings of the IEEE/CVF International Conference on Computer Vision},
  year={2023}
}

@misc{ren2024groundedsamassemblingopenworld,
      title={Grounded SAM: Assembling Open-World Models for Diverse Visual Tasks}, 
      author={Tianhe Ren and Shilong Liu and Ailing Zeng and Jing Lin and Kunchang Li and He Cao and Jiayu Chen and Xinyu Huang and Yukang Chen and Feng Yan and Zhaoyang Zeng and Hao Zhang and Feng Li and Jie Yang and Hongyang Li and Qing Jiang and Lei Zhang},
      year={2024},
      eprint={2401.14159},
      archivePrefix={arXiv},
      primaryClass={cs.CV},
      url={https://arxiv.org/abs/2401.14159}, 
}

@misc{autodistil,
author = {Roboflow},
  title = {Home - Autodistill},
  url = "https://docs.autodistill.com/#license",
month = {},
year = {},
  note = "[Online; accessed 2026-03-04]"
}

@article{xiong2023efficientsam,
  title={EfficientSAM: Leveraged Masked Image Pretraining for Efficient Segment Anything},
  author={Yunyang Xiong and Bala Varadarajan and Lemeng Wu and Xiaoyu Xiang and Fanyi Xiao and Chenchen Zhu and Xiaoliang Dai and Dilin Wang and Fei Sun and Forrest Iandola and Raghuraman Krishnamoorthi and Vikas Chandra},
  journal={arXiv:2312.00863},
  year={2023}
}

@article{dendorfer2021motchallenge,
  title={MOTChallenge: A Benchmark for Single-Camera Multiple Target Tracking: P. Dendorfer et al.},
  author={Dendorfer, Patrick and Osep, Aljosa and Milan, Anton and Schindler, Konrad and Cremers, Daniel and Reid, Ian and Roth, Stefan and Leal-Taix{\'e}, Laura},
  journal={International Journal of Computer Vision},
  volume={129},
  number={4},
  pages={845--881},
  year={2021},
  publisher={Springer}
}

@article{awais2025foundation,
  title={Foundation models defining a new era in vision: a survey and outlook},
  author={Awais, Muhammad and Naseer, Muzammal and Khan, Salman and Anwer, Rao Muhammad and Cholakkal, Hisham and Shah, Mubarak and Yang, Ming-Hsuan and Khan, Fahad Shahbaz},
  journal={IEEE Transactions on Pattern Analysis and Machine Intelligence},
  volume={47},
  number={4},
  pages={2245--2264},
  year={2025},
  publisher={IEEE}
}

@InProceedings{Ong_2023_ICCV,
    author    = {Ong, Kian Eng and Ng, Xun Long and Li, Yanchao and Ai, Wenjie and Zhao, Kuangyi and Yeo, Si Yong and Liu, Jun},
    title     = {Chaotic World: A Large and Challenging Benchmark for Human Behavior Understanding in Chaotic Events},
    booktitle = {Proceedings of the IEEE/CVF International Conference on Computer Vision (ICCV)},
    month     = {October},
    year      = {2023},
    pages     = {20213-20223}
}

@inproceedings{he2025learning,
  title={Learning extremely high density crowds as active matters},
  author={He, Feixiang and Yue, Jiangbei and Zhu, Jialin and Seyfried, Armin and Casas, Dan and Pettr{\'e}, Julien and Wang, He},
  booktitle={Proceedings of the IEEE/CVF Conference on Computer Vision and Pattern Recognition},
  pages={540--550},
  year={2025}
}

@article{shao2018crowdhuman,
    title={CrowdHuman: A Benchmark for Detecting Human in a Crowd},
    author={Shao, Shuai and Zhao, Zijian and Li, Boxun and Xiao, Tete and Yu, Gang and Zhang, Xiangyu and Sun, Jian},
    journal={arXiv preprint arXiv:1805.00123},
    year={2018}
  }

@article{zhang2019widerperson,
Author = {Zhang, Shifeng and Xie, Yiliang and Wan, Jun and Xia, Hansheng and Li, Stan Z. and Guo, Guodong},
journal = {IEEE Transactions on Multimedia (TMM)},
Title = {WiderPerson: A Diverse Dataset for Dense Pedestrian Detection in the Wild},
Year = {2019}}

@misc{kirillov2023segment,
      title={Segment Anything}, 
      author={Alexander Kirillov and Eric Mintun and Nikhila Ravi and Hanzi Mao and Chloe Rolland and Laura Gustafson and Tete Xiao and Spencer Whitehead and Alexander C. Berg and Wan-Yen Lo and Piotr Dollár and Ross Girshick},
      year={2023},
      eprint={2304.02643},
      archivePrefix={arXiv},
      primaryClass={cs.CV},
      url={https://arxiv.org/abs/2304.02643}, 
}

@inproceedings{liu2024grounding,
  title={Grounding dino: Marrying dino with grounded pre-training for open-set object detection},
  author={Liu, Shilong and Zeng, Zhaoyang and Ren, Tianhe and Li, Feng and Zhang, Hao and Yang, Jie and Jiang, Qing and Li, Chunyuan and Yang, Jianwei and Su, Hang and others},
  booktitle={European conference on computer vision},
  pages={38--55},
  year={2024},
  organization={Springer}
}

@article{mchugh2012interrater,
  title={Interrater reliability: the kappa statistic},
  author={McHugh, Mary L},
  journal={Biochemia medica},
  volume={22},
  number={3},
  pages={276--282},
  year={2012},
  publisher={Hrvatsko dru{\v{s}}tvo za medicinsku biokemiju i laboratorijsku medicinu}
}

@article{cohen1960coefficient,
  title={A coefficient of agreement for nominal scales},
  author={Cohen, Jacob},
  journal={Educational and psychological measurement},
  volume={20},
  number={1},
  pages={37--46},
  year={1960},
  publisher={Sage Publications Sage CA: Thousand Oaks, CA}
}

@article{sun2025towards,
  title={Towards pedestrian head tracking: A benchmark dataset and a multi-source data fusion network},
  author={Sun, Kailai and Wang, Xinwei and Liu, Shaobo and Zhao, Qianchuan and Huang, Gao and Liu, Chang},
  journal={Engineering Applications of Artificial Intelligence},
  volume={158},
  pages={111265},
  year={2025},
  publisher={Elsevier}
}

@article{li2023hrbust,
  title={HRBUST-LLPED: a benchmark dataset for wearable low-light pedestrian detection},
  author={Li, Tianlin and Sun, Guanglu and Yu, Linsen and Zhou, Kai},
  journal={Micromachines},
  volume={14},
  number={12},
  pages={2164},
  year={2023},
  publisher={MDPI}
}

@misc{dollar_wojek_schiele_perona_2009, title={Caltech Pedestrians}, DOI={10.1109/CVPR.2009.5206631}, abstractNote={The dataset contains richly annotated video, recorded from a moving vehicle, with challenging images of low resolution and frequently occluded people. We propose improved evaluation metrics, demonstrating that commonly used per-window measures are flawed and can fail to predict performance on full images. This zip file includes folders containing the dataset, code, files, and ROC curve results.}, publisher={IEEE Conference on Computer Vision and Pattern Recognition}, author={Dollar, Piotr and Wojek, Christian and Schiele, Bernt and Perona, Pietro}, year={2009}, month={Jun} }

@inproceedings{zhang2017citypersons,
  title={Citypersons: A diverse dataset for pedestrian detection},
  author={Zhang, Shanshan and Benenson, Rodrigo and Schiele, Bernt},
  booktitle={Proceedings of the IEEE conference on computer vision and pattern recognition},
  pages={3213--3221},
  year={2017}
}

@inproceedings{sun2022dance,
    title={DanceTrack: Multi-Object Tracking in Uniform Appearance and Diverse Motion},
    author={Sun, Peize and Cao, Jinkun and Jiang, Yi and Yuan, Zehuan and Bai, Song and Kitani, Kris and Luo, Ping},
    booktitle={Proceedings of the IEEE/CVF Conference on Computer Vision and Pattern Recognition (CVPR)},
    year={2022}
}

@article{leal2015motchallenge,
  title={Motchallenge 2015: Towards a benchmark for multi-target tracking},
  author={Leal-Taix{\'e}, Laura and Milan, Anton and Reid, Ian and Roth, Stefan and Schindler, Konrad},
  journal={arXiv preprint arXiv:1504.01942},
  year={2015}
}

@misc{milan2016mot16benchmarkmultiobjecttracking,
      title={MOT16: A Benchmark for Multi-Object Tracking}, 
      author={Anton Milan and Laura Leal-Taixe and Ian Reid and Stefan Roth and Konrad Schindler},
      year={2016},
      eprint={1603.00831},
      archivePrefix={arXiv},
      primaryClass={cs.CV},
      url={https://arxiv.org/abs/1603.00831}, 
}

@misc{dendorfer2020mot20benchmarkmultiobject,
      title={MOT20: A benchmark for multi object tracking in crowded scenes}, 
      author={Patrick Dendorfer and Hamid Rezatofighi and Anton Milan and Javen Shi and Daniel Cremers and Ian Reid and Stefan Roth and Konrad Schindler and Laura Leal-Taixé},
      year={2020},
      eprint={2003.09003},
      archivePrefix={arXiv},
      primaryClass={cs.CV},
      url={https://arxiv.org/abs/2003.09003}, 
}

@article{du2024exploring,
  title={Exploring the state-of-the-art in multi-object tracking: A comprehensive survey, evaluation, challenges, and future directions},
  author={Du, Chenjie and Lin, Chenwei and Jin, Ran and Chai, Bencheng and Yao, Yingbiao and Su, Siyu},
  journal={Multimedia tools and applications},
  volume={83},
  number={29},
  pages={73151--73189},
  year={2024},
  publisher={Springer}
}

@article{zhang2022dino,
  title={Dino: Detr with improved denoising anchor boxes for end-to-end object detection},
  author={Zhang, Hao and Li, Feng and Liu, Shilong and Zhang, Lei and Su, Hang and Zhu, Jun and Ni, Lionel M and Shum, Heung-Yeung},
  journal={arXiv preprint arXiv:2203.03605},
  year={2022}
}

@article{ren2024grounded,
  title={Grounded sam: Assembling open-world models for diverse visual tasks},
  author={Ren, Tianhe and Liu, Shilong and Zeng, Ailing and Lin, Jing and Li, Kunchang and Cao, He and Chen, Jiayu and Huang, Xinyu and Chen, Yukang and Yan, Feng and others},
  journal={arXiv preprint arXiv:2401.14159},
  year={2024}
}

@article{landis1977application,
  title={An application of hierarchical kappa-type statistics in the assessment of majority agreement among multiple observers},
  author={Landis, J Richard and Koch, Gary G},
  journal={Biometrics},
  pages={363--374},
  year={1977},
  publisher={JSTOR}
}

@inproceedings{zhang2024when,
  title={When Pedestrian Detection Meets Multi-Modal Learning: Generalist Model and Benchmark Dataset},
  author={Zhang, Yi and Zeng, Wang and Jin, Sheng and Qian, Chen and Luo, Ping and Liu, Wentao},
  booktitle={European Conference on Computer Vision (ECCV)},
  year={2024},
  month={September}
}

@inproceedings{fabbri2018learning,
   title     = {Learning to Detect and Track Visible and Occluded Body Joints in a Virtual World},
   author    = {Fabbri, Matteo and Lanzi, Fabio and Calderara, Simone and Palazzi, Andrea and Vezzani, Roberto and Cucchiara, Rita},
   booktitle = {European Conference on Computer Vision (ECCV)},
   year      = {2018}
 }

@inproceedings{fabbri2021motsynth,
  title={Motsynth: How can synthetic data help pedestrian detection and tracking?},
  author={Fabbri, Matteo and Bras{\'o}, Guillem and Maugeri, Gianluca and Cetintas, Orcun and Gasparini, Riccardo and O{\v{s}}ep, Aljo{\v{s}}a and Calderara, Simone and Leal-Taix{\'e}, Laura and Cucchiara, Rita},
  booktitle={Proceedings of the IEEE/CVF international conference on computer vision},
  pages={10849--10859},
  year={2021}
}

@inproceedings{le2024jrdb,
  title={Jrdb-panotrack: An open-world panoptic segmentation and tracking robotic dataset in crowded human environments},
  author={Le, Duy Tho and Gou, Chenhui and Datta, Stavya and Shi, Hengcan and Reid, Ian and Cai, Jianfei and Rezatofighi, Hamid},
  booktitle={Proceedings of the IEEE/CVF Conference on Computer Vision and Pattern Recognition},
  pages={22325--22334},
  year={2024}
}

@article{HUANG2023106285,
title = {A machine vision-based method for crowd density estimation and evacuation simulation},
journal = {Safety Science},
volume = {167},
pages = {106285},
year = {2023},
issn = {0925-7535},
doi = {https://doi.org/10.1016/j.ssci.2023.106285},
url = {https://www.sciencedirect.com/science/article/pii/S0925753523002278},
author = {Shijie Huang and Jingwei Ji and Yu Wang and Wenju Li and Yuechuan Zheng}
}

@inproceedings{chen2019crowd,
  title={Crowd-robot interaction: Crowd-aware robot navigation with attention-based deep reinforcement learning},
  author={Chen, Changan and Liu, Yuejiang and Kreiss, Sven and Alahi, Alexandre},
  booktitle={2019 international conference on robotics and automation (ICRA)},
  pages={6015--6022},
  year={2019},
  organization={IEEE}
}

@article{aharon2022bot,
  title={BoT-SORT: Robust associations multi-pedestrian tracking},
  author={Aharon, Nir and Orfaig, Roy and Bobrovsky, Ben-Zion},
  journal={arXiv preprint arXiv:2206.14651},
  year={2022}
}

@inproceedings{cao2023observation,
  title={Observation-centric sort: Rethinking sort for robust multi-object tracking},
  author={Cao, Jinkun and Pang, Jiangmiao and Weng, Xinshuo and Khirodkar, Rawal and Kitani, Kris},
  booktitle={Proceedings of the IEEE/CVF conference on computer vision and pattern recognition},
  pages={9686--9696},
  year={2023}
}

@inproceedings{zhang2022bytetrack,
  title={Bytetrack: Multi-object tracking by associating every detection box},
  author={Zhang, Yifu and Sun, Peize and Jiang, Yi and Yu, Dongdong and Weng, Fucheng and Yuan, Zehuan and Luo, Ping and Liu, Wenyu and Wang, Xinggang},
  booktitle={European conference on computer vision},
  pages={1--21},
  year={2022},
  organization={Springer}
}

@article{sapkota2025yolo26,
  title={YOLO26: key architectural enhancements and performance benchmarking for real-time object detection},
  author={Sapkota, Ranjan and Cheppally, Rahul Harsha and Sharda, Ajay and Karkee, Manoj},
  journal={arXiv preprint arXiv:2509.25164},
  year={2025}
}

@inproceedings{sohan2024review,
  title={A review on yolov8 and its advancements},
  author={Sohan, Mupparaju and Sai Ram, Thotakura and Rami Reddy, Ch Venkata},
  booktitle={International conference on data intelligence and cognitive informatics},
  pages={529--545},
  year={2024},
  organization={Springer}
}

@inproceedings{zhao2024detrs,
  title={Detrs beat yolos on real-time object detection},
  author={Zhao, Yian and Lv, Wenyu and Xu, Shangliang and Wei, Jinman and Wang, Guanzhong and Dang, Qingqing and Liu, Yi and Chen, Jie},
  booktitle={Proceedings of the IEEE/CVF conference on computer vision and pattern recognition},
  pages={16965--16974},
  year={2024}
}

@article{sapkota2025ultralytics,
  title={Ultralytics YOLO evolution: An overview of YOLO26, YOLO11, YOLOv8 and YOLOv5 object detectors for computer vision and pattern recognition},
  author={Sapkota, Ranjan and Karkee, Manoj},
  journal={arXiv preprint arXiv:2510.09653},
  year={2025}
}

@article{vijayakumar2024yolo,
  title={Yolo-based object detection models: A review and its applications},
  author={Vijayakumar, Ajantha and Vairavasundaram, Subramaniyaswamy},
  journal={Multimedia Tools and Applications},
  volume={83},
  number={35},
  pages={83535--83574},
  year={2024},
  publisher={Springer}
}

@inproceedings{barsellotti2025talking,
  title={Talking to dino: Bridging self-supervised vision backbones with language for open-vocabulary segmentation},
  author={Barsellotti, Luca and Bianchi, Lorenzo and Messina, Nicola and Carrara, Fabio and Cornia, Marcella and Baraldi, Lorenzo and Falchi, Fabrizio and Cucchiara, Rita},
  booktitle={Proceedings of the IEEE/CVF International Conference on Computer Vision},
  pages={22025--22035},
  year={2025}
}

@article{zhang2025efficiently,
  title={How to efficiently annotate images for best-performing deep learning-based segmentation models: An empirical study with weak and noisy annotations and segment anything model},
  author={Zhang, Yixin and Zhao, Shen and Gu, Hanxue and Mazurowski, Maciej A},
  journal={Journal of Imaging Informatics in Medicine},
  volume={38},
  number={5},
  pages={3235--3247},
  year={2025},
  publisher={Springer}
}

@inproceedings{tschirschwitz2025label,
  title={Label convergence: Defining an upper performance bound in object recognition through contradictory annotations},
  author={Tschirschwitz, David Eike and Rodehorst, Volker},
  booktitle={Proceedings of the Winter Conference on Applications of Computer Vision},
  pages={6848--6857},
  year={2025}
}

@inproceedings{ma2022effect,
  title={The effect of improving annotation quality on object detection datasets: A preliminary study},
  author={Ma, Jiaxin and Ushiku, Yoshitaka and Sagara, Miori},
  booktitle={Proceedings Of The IEEE/CVF conference on computer vision and pattern recognition},
  pages={4850--4859},
  year={2022}
}

@inproceedings{alhazmi2021effects,
  title={Effects of annotation quality on model performance},
  author={Alhazmi, Khaled and Alsumari, Walaa and Seppo, Indrek and Podkuiko, Lara and Simon, Martin},
  booktitle={2021 international conference on artificial intelligence in information and communication (ICAIIC)},
  pages={063--067},
  year={2021},
  organization={IEEE}
}

@misc{nae2026learningflyreplaybasedcontinual,
      title={Learning on the Fly: Replay-Based Continual Object Perception for Indoor Drones}, 
      author={Sebastian-Ion Nae and Mihai-Eugen Barbu and Sebastian Mocanu and Marius Leordeanu},
      year={2026},
      eprint={2602.13440},
      archivePrefix={arXiv},
      primaryClass={cs.CV},
      url={https://arxiv.org/abs/2602.13440}, 
}
}

\end{document}